\documentclass[runningheads]{llncs}
\usepackage{graphicx}
\usepackage{amsmath}
\usepackage{booktabs}
\usepackage{algorithm}
\usepackage{algorithmic}
\usepackage{amssymb}
\usepackage{amsfonts}
\usepackage{textcomp}
\usepackage{xcolor}
\usepackage{bm}
\usepackage{dsfont}
\usepackage{centernot}
\usepackage{multirow}
\usepackage{caption}
\usepackage{subcaption}
\begin{document}
\title{ARES: Locally Adaptive Reconstruction-based Anomaly Scoring} 

\author{Adam Goodge \inst{1,3}\and
Bryan Hooi \inst{1,2} \and
See Kiong Ng \inst{1,2} \and
Wee Siong Ng \inst{3}}

\authorrunning{A. Goodge et al.}


\institute{School of Computing, National University of Singapore \and
           Institute of Data Science, National University of Singapore \and
           Institute for Infocomm Research, A*STAR, Singapore \\
\email{adam.goodge@u.nus.edu,\{dcsbhk, seekiong\}@nus.edu.sg, wsng@i2r.a-star.edu.sg}}
\maketitle              

\begin{abstract}
How can we detect anomalies: that is, samples that significantly differ from a given set of high-dimensional data, such as images or sensor data? This is a practical problem with numerous applications and is also relevant to the goal of making learning algorithms more robust to unexpected inputs. Autoencoders are a popular approach, partly due to their simplicity and their ability to perform dimension reduction. However, the anomaly scoring function is not adaptive to the natural variation in reconstruction error across the range of normal samples, which hinders their ability to detect real anomalies. In this paper, we empirically demonstrate the importance of local adaptivity for anomaly scoring in experiments with real data. We then propose our novel Adaptive Reconstruction Error-based Scoring approach, which adapts its scoring based on the local behaviour of reconstruction error over the latent space. We show that this improves anomaly detection performance over relevant baselines in a wide variety of benchmark datasets.
\end{abstract}

\keywords{Anomaly detection \and machine learning \and unsupervised learning.}

\section{Introduction}

The detection of anomalous data is important in a wide variety of applications, such as detecting fraudulent financial transactions or malignant tumours. Recently, deep learning methods have enabled significant improvements in performance on ever larger and higher-dimensional datasets. Despite this, anomaly detection remains a challenging task, most notably due to the difficulty of obtaining accurate labels of anomalous data. For this reason, supervised classification methods are unsuitable for anomaly detection. Instead, unsupervised methods are used to learn the distribution of an all-normal training set. Anomalies are then detected amongst unseen samples by measuring their closeness to the normal-data distribution.

Autoencoders are an extremely popular approach to learn the behaviour of normal data. The reconstruction error of a sample is directly used as its anomaly score; anomalies are assumed to have a higher reconstruction error than normal samples due to their difference in distribution. However, this anomaly score fails to account for the fact that reconstruction error can vary greatly even amongst different types of normal samples. For example, consider a sensor system in a factory with various activities on weekdays but zero activity on weekends. The weekday samples are diverse and complex, therefore we could expect the reconstruction error of these samples to vary greatly. Meanwhile, even a small amount of activity in a weekend sample would be anomalous, even if the effect on the reconstruction error is minimal. The anomaly detector would likely detect false positives (high reconstruction error) among weekday samples and false negatives among weekend samples. Although encoded within the input data attributes, the context of each individual sample (i.e. day of the week) is neglected at the detection stage by the standard scoring approach. Instead, all samples are assessed according to the same error threshold or standard. This invokes the implicit assumption that the reconstruction errors of all samples are identically distributed, regardless of any individual characteristics and context which could potentially influence the reconstruction error significantly.

In this work, we aim to address this problem by proposing \textbf{Adaptive Reconstruction Error-based Scoring (\textsc{ARES})}. Our \textbf{locally-adaptive} scoring method is able to automatically account for any contextual information which affects the reconstruction error, resulting in more accurate anomaly detection. We use a flexible, \textbf{neighbourhood-based} approach to define the context, based on the location of its latent representation learnt by the model.

Our scoring approach is applied at test time, so it can be retrofitted to pre-trained models of any size and architecture or used to complement existing anomaly detection techniques. Our score is simple and efficient to compute, requiring very little additional computational time, and the code is available online\footnote{https://github.com/agoodge/ARES}. In summary, we make the following contributions:
\begin{enumerate}
    \item We \textbf{empirically show} with \textbf{real, multi-class data} the variation in reconstruction error among different samples in the normal set and why this justifies the need for local adaptivity in anomaly scoring.
    
    \item We propose a novel anomaly scoring method which adapts to this variation by evaluating anomaly status based on the local context of a given sample in the latent space.
    
    \item We evaluate our method against a wide range of baselines with various benchmark datasets. We also study the effect of different components and formulations of our scoring method in an ablation study.
    
\end{enumerate}

\section{Background}\label{background}

\subsection{Autoencoders}

Autoencoders are neural networks that output a reconstruction of the input data \cite{Sakurada2014}. They are comprised of two components: an encoder and decoder. The encoder compresses data from the input level into lower-dimensional latent representations through its hidden layers. The encoder output is typically known as the bottleneck, from which the reconstruction of the original data is found through the decoder hidden layers. The network is trained to minimise the reconstruction error over the training set:
\begin{align}
\min_{\theta, \phi} \Vert \mathbf{x} - (f_\theta \circ g_\phi) (\mathbf{x})\Vert_2^2,
\end{align}
where $g_\phi$ is the encoder, $f_\theta$ the decoder. The assumption is that data lies on a lower-dimensional manifold within the high-dimensional input space. The autoencoder learns to reconstruct data on this manifold by performing dimension reduction. By training the model on only normal data, the reconstruction error of a normal sample should be low as it close to the learnt manifold on which it has been reconstructed by the autoencoder, whereas anomalies are far away and are reconstructed with higher error.

\subsection{Local Outlier Factor}\label{lof}

The Local Outlier Factor (\textsc{LOF}) method is a neighbourhood-based approach to anomaly detection; it measures the density of a given sample relative to its $k$ nearest neighbours' \cite{Breunig2000}. Anomalies are assumed to be in sparse regions far away from the one or more high-density clusters of normal data. A lower density therefore suggests the sample is anomalous. The original method uses the `reachability distance'; defined for a point $A$ from point $B$ as:
\begin{equation}
\max \{k \text{-distance}(B), d(A,B)\}
\end{equation}
where $d(\cdot,\cdot)$ is a chosen distance metric and $k$-distance$(B)$ is the distance of $B$ to its $k^{th}$ nearest neighbour. The local reachability density and subsequently the local outlier factor of $A$ based on its set of neighbours $\mathsf{N}_{k}(A)$, is:

\begin{equation}
\text{lrd}_k (A) := \left(\frac{\sum_{B\in \mathsf{N}_{k}(A)} \text{reachability-distance}_{k}(A,B)}{|\mathsf{N}_{k}(A)|}\right)^{-1}.
\end{equation}

\begin{equation}\label{lof_score}
\text{LOF}_{k}(A) = \frac{\sum_{B\in \mathsf{N}_{k}(A)} \text{lrd}_k (B)}{|\mathsf{N}_{k}(A)| \cdot \text{lrd}_k (A)},
\end{equation}

\section{Related Work}\label{related_work}

Anomalies are often assumed to occupy sparse regions far away from high-density clusters of normal data. As such, a great variety of methods detect anomalies by measuring the distances to their nearest neighbours, most notably \textsc{KNN} \cite{Zimek2013}, which directly uses the distance of a point to its $k^{th}$ nearest neighbour as the anomaly score. \textsc{LOF} \cite{Breunig2000} measures the density of a point relative to the density of points in its local neighbourhood, as seen in Section \ref{background}. This is beneficial as a given density may be anomalous in one region but normal in another region. This local view accommodates the natural variation in density and therefore allows for the detection of more meaningful anomalies. More recently, deep models such as graph neural networks have been used to learn anomaly scores from neighbour-distances in a data-driven way \cite{goodge2021lunar}. Naturally, this relies on an effective distance metric. This is often itself a difficult problem; even the most established metrics have been shown to lose significance in high-dimensional spaces \cite{Beyer1999} due to the `curse of dimensionality'.

Reconstruction-based methods rely on deep models, such as autoencoders, to learn to reconstruct samples from the normal set accurately. Samples are then flagged as anomalous if their reconstruction error higher than some pre-defined threshold, as it is assumed that the model will reconstruct samples outside of the normal set with a higher reconstruction error \cite{Sakurada2014,Feng2015,An2015,Chen2017,Goodge}. 

More recent works have used autoencoders or other models are deep feature extractors, with the learnt features then used in another anomaly detection module downstream, such as Gaussian mixture models \cite{Bo2018}, DBSCAN \cite{amarbayasgalan2018unsupervised}, KNN \cite{bergman2020deep} and auto-regressive models \cite{Abati2019}. \cite{Deng2022} calibrates different types of autoencoders against the effects of varying hardness within normal samples.

Other methods measure the distance of a sample to a normal set-enclosing hypersphere \cite{tax2004support,Ruff2018}. or its likelihood under a learnt model \cite{dinh2016density,Rezende2015,Zhai2016}. Generative adversarial networks have also been proposed for anomaly detection, mostly relying either on the discriminator score or the accuracy of the generator for a given sample to determine anomalousness \cite{Akcay2018,Zenati2019}.

In all of these methods, the normal set is often restricted to a subset of the available data; e.g. just one class from a multi-class dataset in experiments. In practice, normal data could belong to multiple classes or modes of behaviour which all need to be modelled adequately. As such, developing anomaly detection methods that can model a diverse range of normal behaviours and adapt their scoring appropriately are important.

\section{Methodology}

\subsection{Problem Definition}

We assume to have $m$ normal training samples $\mathbf{x}_1^{\text{(train)}},  \cdots, \mathbf{x}_m^{\text{(train)}} \in \mathbb{R}^d$ and $n$ testing samples, $\mathbf{x}_1^{\text{(test)}},  \cdots, \mathbf{x}_n^{\text{(test)}} \in \mathbb{R}^d$, each of which may be normal or anomalous. For each test sample $\mathbf{x}$, our algorithm should indicate how anomalous it is through computing its \textbf{anomaly score} $s(\mathbf{x})$. The goal is for anomalies to be given higher anomaly scores than normal points. In this work, the fundamental question is:

\textit{Given an autoencoder with encoder $g_\phi$ and decoder $f_\theta$, how can we use the latent encoding $\mathbf{z} = g_\phi(\mathbf{x}$) and reconstruction $\hat{\mathbf{x}} = (f_\theta \circ g_\phi)(\mathbf{x})$ of a sample $\mathbf{x}$ to score its anomalousness?} 

Our approach is not limited to standard autoencoders, and can be applied to other models that involve a latent encoding $\mathbf{z}$ and a reconstruction $\hat{\mathbf{x}}$.

In practice, the anomaly score is compared with a user-defined anomaly threshold; samples which exceed this threshold are flagged as anomalies. Different approaches can be employed to set this threshold, such as extreme value theory~\cite{siffer2017anomaly}. Our focus is instead on the approach to anomaly scoring itself, which allows for any choice of thresholding scheme to be used alongside it. 

\subsection{Statistical Interpretation of Reconstruction-based Anomaly Detection}\label{statistical_interpretation}

In the standard approach, with the residual as $\bm{\varepsilon} := \mathbf{x} - \hat{\mathbf{x}}$, the anomaly score is:
\begin{equation}
    R(\bm{\varepsilon}) := \Vert \bm{\varepsilon} \Vert_2^2 = \Vert \mathbf{x} - \hat{\mathbf{x}} \Vert_2^2.
\end{equation}
This can be seen as a negative log-likelihood-based score that assumes $\bm{\varepsilon}$ follows a Gaussian distribution with zero mean and unit variance: $\bm{\varepsilon} \sim \mathcal{N}(0, \mathbf{I})$:
\begin{equation}
-\log P(\bm{\varepsilon}) = \frac{d}{2}\log(2\pi) + \frac{R(\bm{\varepsilon})}{2}
\end{equation}
The negative log-likelihood is an intuitive anomaly score because anomalous samples should have lower likelihood, thus higher negative log-likelihood.

The key conclusion, ignoring the constant additive and multiplicative factors, is that $R(\bm{\varepsilon})$ can be equivalently seen as a negative log-likelihood-based score, based on a model with the implicit assumption of constant mean as well as \emph{constant variance residuals across all samples}, known in the statistical literature as \textbf{homoscedasticity}. Most crucially, this assumption applies to all $\mathbf{x}$ regardless of its individual latent representation, $\mathbf{z}$, which implies that the reconstruction error of $\mathbf{x}$ does not depend on the location of $\mathbf{z}$ in the latent space. We question the validity of this assumption in real datasets.

In our earlier example of a factory sensor system, the variance of a sample $\mathbf{x}$ depends greatly on its characteristics (i.e. day of the week), with high variance during weekdays and low variance on weekends. This is instead a prime example of \textbf{heteroscedasticity}. Furthermore, since the latent representation $\mathbf{z}$ typically encodes these important characteristics of $\mathbf{x}$, there is likely to be a clear dependence between $\mathbf{z}$ and the reconstruction error $R(\bm{\varepsilon})$, which will be examined in Section \ref{motivation}. This relationship could be exploited to improve the modelling of reconstruction errors and thus detection accuracy by adapting the anomaly score to this relationship at the sample-level.

\subsection{Motivation \& Empirical Results}\label{motivation}

\begin{figure}
    \centering
    \includegraphics[width=0.7\linewidth]{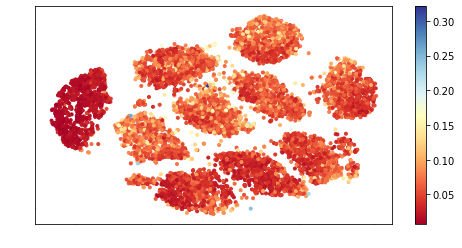}
    \caption{t-SNE plot of the latent encodings with colour determined by reconstruction error of the associated sample.}
    \label{M2}
\end{figure}

In this section, we train an autoencoder on all 10 classes of \texttt{MNIST} to empirically examine the variation in the reconstruction error of real data to scrutinise the homoscedastic assumption. In doing so, we exemplify the shortcomings of the standard approach which implicitly assumes all reconstruction errors from a fixed Gaussian distribution. The autoencoder architecture and training procedure is the same as those described in Section \ref{sec:experiments}. We make the following observations:

\begin{enumerate}
    \item \textbf{Inter-class variation:} In Figure \ref{M2}, the t-SNE \cite{tsne} projections of the latent representations of training points are coloured according to their reconstruction error. We see that the autoencoder learns to separate each class approximately into distinct clusters. In this context, the class label can be seen as a variable characteristic between different samples within the normal set (similar to the weekday vs. weekend example). 
    
    \begin{figure}
    \centering
        \includegraphics[width=0.7\linewidth]{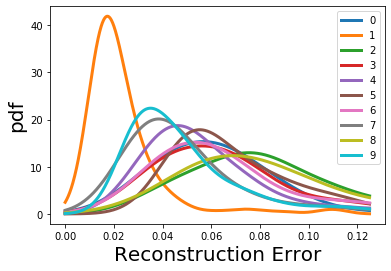}
    \caption{Probability density function of reconstruction errors associated with different classes of training samples, estimated via kernel density estimation.}
    \label{M1}
\end{figure}
    
    \item \textbf{Intra-class variation} Also in Figure \ref{M2}, we can see that there is significant variation even within any single cluster. In other words, there are noticeably distinct regions of high (and low) reconstruction errors within each individual cluster. This shows that there are additional characteristics that influence whether a given normal sample has a high or low reconstruction error besides its class label.
    
    We observe that there is significant variation in reconstruction error between the different clusters. Most notably, samples in the leftmost cluster (corresponding to class 1) has significantly lower reconstruction errors than most others. Indeed, as shown in Figure \ref{M1}, the distribution of reconstruction errors associated with class 1 samples is very different to those of the other classes. 
    
    This shows that there is significant variation in reconstruction errors between classes, and that it is inappropriate to assume that the reconstruction errors of all samples can be modelled by a single, fixed Gaussian distribution. For example, a reconstruction error of $0.06$ would be high for a class 1 or class 9 sample, but low for a class 2 or class 8 sample. 
    
    \begin{figure}
\centering
    \includegraphics[width=0.7\linewidth]{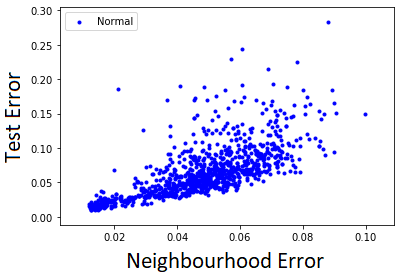}
    \caption{Average reconstruction error of the training set nearest neighbours of a point versus its own reconstruction error for normal test samples.}
    \label{M3}
\end{figure}
    
    \item \textbf{Neighbourhood correlation:} In Figure \ref{M3}, we plot the reconstruction error of test samples against the average reconstruction error of its nearest training set neighbours in the latent space (neighbourhood error). We see that the reconstruction error of a test point increases as its neighbourhood error increases. Furthermore, the variance in test errors increases for larger neighbourhood errors: a clear heteroscedastic relationship. This information is useful in determining anomalousness: a test error of $0.1$ would be anomalously high if its neighbourhood error is $0.02$, but normal if it is $0.06$.
\end{enumerate}
Given these observations, we propose that anomalies could be more accurately detected by incorporating contextual information into the anomaly scoring beyond reconstruction error alone. Furthermore, analysing the neighbourhood of a given sample provides this contextual information to help determine its anomalousness. In section \ref{subsec:ARES}, we propose our Adaptive Reconstruction Error-based Scoring method to achieve this.

\subsection{Adaptive Reconstruction Error-based Scoring}\label{subsec:ARES}

In Section \ref{statistical_interpretation}, we saw that the standard approach assumes that all residuals come from a fixed Gaussian with constant mean and unit variance: $\bm{\varepsilon} \sim \mathcal{N}(0, \mathbf{I})$. In Section \ref{motivation}, we saw that this assumption is inappropriate. In this section, we detail \textsc{ARES}, a novel anomaly scoring methodology which aims to address this flaw by adapting the scoring for each samples local context in the latent space.

The normal level of reconstruction error varies for samples in different regions of the latent space, meaning the latent encoding of a sample holds important information regarding anomalousness. As such, \textsc{ARES} is inspired by the \textbf{joint-likelihood} of a samples residual $\bm{\varepsilon}$ with its latent encoding $\mathbf{z}$, defined as follows: 

\begin{equation}\label{probability}
-\log P(\bm{\varepsilon},\mathbf{z}) =  -\log P(\bm{\varepsilon}|\mathbf{z}) -\log P(\mathbf{z}).
\end{equation}

The first term, $-\log P(\bm{\varepsilon}|\mathbf{z})$, is the conditional (negative log-) likelihood of the points residual conditioned on its latent encoding. The second term, $-\log P(\mathbf{z})$, is the likelihood of observing the latent encoding from the normal set.

We now detail our approach to interpret these terms into tractable, efficient scores which we name the \textbf{local reconstruction score} and \textbf{local density score} respectively. These scores are combined to give the overall \textsc{ARES} anomaly score.

\subsection{Local Reconstruction Score}

The local reconstruction score is based on an estimate of how likely a given residual $\bm{\varepsilon}$ (and consequent reconstruction error) is to come from the corresponding sample $\mathbf{x}$, based on its latent encoding $\mathbf{z}$:

\begin{equation}
    r(\mathbf{x}) = -\log P(\bm{\varepsilon}|\mathbf{z}) 
\end{equation}

This likelihood cannot be calculated directly for any individual $\mathbf{z}$. Instead, we consider the $k$ nearest neighbours of $\mathbf{z}$ in the latent space, denoted $\mathsf{N}_k(\mathbf{z})$, to be a sample population which $z$ belongs to. This is intuitive as data points with similar characteristics to $\textbf{x}$ in the input space are more likely to be encoded nearer to $\mathbf{z}$ in the latent space. The full local reconstruction score algorithm is shown in Algorithm \ref{ldrs}.

After training the autoencoder, we fix the model weights and store in memory all training samples' reconstruction errors and latent encodings taken from the bottleneck layer. For a test sample $\mathbf{x}$, we find its own latent encoding $\mathbf{z}$ and reconstruction $\hat{\mathbf{x}}$. We then find $\mathsf{N}_k(\mathbf{z})$, the set of $k$ nearest neighbours to $\mathbf{z}$ amongst the training set encodings. We only find nearest neighbours among the training samples as they are all assumed to be normal and therefore should have reconstruction errors within the normal range.

We want to obtain an estimate of the reconstruction error that could be expected of $\mathbf{x}$, assuming its a normal point, based on the reconstruction errors of its (normal) neighbours. We can then compare this expected value, conditioned on its unique location in the latent space, with the points true reconstruction error to determine its anomalousness.

In a fully probabilistic approach, we could do this by measuring the likelihood of the test points reconstruction error under a probability distribution, e.g. a Gaussian, fit to the neighbours' reconstruction errors. However, it is unnecessarily restrictive to assume any closed-form probability distribution to adequately model this population for the unique neighbourhood of each and every test sample. Instead, we opt for a non-parametric approach; we measure the difference between the test points reconstruction error and the median reconstruction error of its neighbouring samples. The larger the difference between them, the more outlying the test point is in comparison to its local neighbours, therefore the more likely it is to be anomalous. In practice, using the median was found to perform better than the mean as it is more robust to extrema. With this, we obtain the local reconstruction score as: 
\begin{equation}\label{eq:lrd}
  r(\mathbf{x}) = \Vert \mathbf{x} - \mathbf{\hat{x}} \Vert_2^2  - \underset{\mathbf{n} \in \mathsf{N}_k(\mathbf{z})}{\mathsf{median}}(\Vert \mathbf{n} - \hat{\mathbf{n}} \Vert_2^2).
\end{equation}

Nearest neighbour search in the latent space is preferable over the input space as dimension reduction helps to alleviates the curse of dimensionality (see Section \ref{related_work}), resulting in more semantically-meaningful neighbors. Secondly, from a practical perspective, the neighbour search is less time-consuming and computationally intensive in lower dimensional spaces.

\begin{algorithm}
\caption{Local Reconstruction Score}
\label{ldrs}
\textbf{Input}: Autoencoder $A(\cdot) = (f_\theta \circ g_\phi )(\cdot)$, training set $\mathbf{X}_{\text{train}}$, test sample $\mathbf{x}_{\text{test}}$\\
\textbf{Parameters}: neighbour count $k$\\
\textbf{Output}: Local reconstruction score $r(\mathbf{x}_{\text{test}})$ \\
\begin{algorithmic}[1] 

\STATE Train autoencoder $A$ on training set according to: $\min_{\theta, \phi} \Vert \mathbf{X}_{\text{train}} - \mathbf{\hat{X}}_{\text{train}} \Vert _2^2$ where $\mathbf{\hat{x}}^{i}_{\text{train}} = (f_\theta \circ g_\phi )(\mathbf{x}_{\text{train}}^{i})$ for 
$\mathbf{x}_{\text{train}}^{i} \in \mathbf{X}_{\text{train}}$.
\\

\STATE Find latent encoding of $\mathbf{x}_{\text{test}}$: $\mathbf{z}_{\text{test}} := g_\phi(\mathbf{x}_{\text{test}})$\\

\STATE Find the set of $k$ nearest neighbours to $\mathbf{z}_{\text{test}}$ among latent encodings of the training data: $\mathsf{N}_k(\mathbf{z}_{\text{test}}) := \{\mathbf{z}_{\text{train}}^{1},...,\mathbf{z}_{\text{train}}^{k}\}$ where  $\mathbf{z}_{\text{train}}^{i} = g_\phi(\mathbf{x}^{i}_{\text{train}}) \in$  for $i = \{1,...,k\}$. \\ 

\STATE \textbf{return} $r(\mathbf{x}_{\text{test}}) = \Vert \mathbf{x}_{\text{test}} - \mathbf{\hat{x}}_{\text{test}} \Vert_2^2  - \underset{\mathbf{n} \in \mathsf{N}_k(\mathbf{z})}{\mathsf{median}}(\Vert \mathbf{x}^{i}_{\text{train}}- \mathbf{\hat{x}}^{i}_{\text{train}} \Vert_2^2)$
\end{algorithmic}
\end{algorithm}

\subsection{Local Density Score}\label{density_score}

The local reconstruction score corresponds to the conditional term of the joint likelihood. We now introduce the local density score, which corresponds to the likelihood of observing the given encoding in the latent space. This is a density estimation task, which concerns the relative distance of $\mathbf{z}$ to its nearest neighbours, unlike the local reconstruction score which focuses on the reconstruction error of neighbours. Anomalies are assumed to exist in sparse regions, where normal samples are unlikely to be found in significant numbers. Any multivariate distribution $P$, with trainable parameters $\Theta$, could be used to estimate this density:

\begin{align}
d(\mathbf{x}) := -\log P(\mathbf{z};\Theta) \label{eq:d}
\end{align}

Note that it is common to ignore constant factor shifts in the anomaly score. Thus, the distribution $P$ need not be normalized; even unnormalized density estimation techniques can be used as scoring functions. We note that \textsc{LOF} is an example of an unnormalized score which is similarly locally adaptive like the local reconstruction score, so \textsc{LOF} is used for the local density score in our main experiments. Other methods are also tested and their performance is shown in the ablation study. 

The overall anomaly score for sample $\mathbf{x}$ is: 
\begin{equation}\label{anomaly_score}
s(\mathbf{x}) := r(\mathbf{x}) + \alpha d(\mathbf{x}),
\end{equation}
where $r(\mathbf{x})$ is its local reconstruction score and $d(\mathbf{x})$ the local density score. These two scores are unnormalized, so we use a scaling factor $\alpha$ to balance the relative magnitudes of the two scores. We heuristically set it equal to $0.5$ for all datasets and settings for simplicity, as this was found to balance the two scores sufficiently fairly in most cases. We choose not to treat $\alpha$ as a hyper-parameter to be tuned to optimise performance, although different values could give better performance for different datasets. The effect of changing $\alpha$ is shown in the supplementary material.

\paragraph{Computational Runtime:} The average runtimes of experiments with the \texttt{MNIST} dataset can be found in the supplementary material. We see that, despite taking longer than the standard reconstruction error approach, the additional computational runtime of \textsc{ARES} anomaly scoring is insignificant in relation to the model training time. Anomaly scoring with \textsc{ARES} is just $1.2\%$ of the overall time taken (including training) in the one-class case (0.017 minutes for scoring versus 1.474 minutes for training), and $3.5\%$ in the multi-class case. The additional run-time is a result of the $k$ nearest neighbour search. An exact search algorithms would be $\mathcal{O}(nm)$ with $n$ train and $m$ test samples, however approximate methods can achieve near-exact accuracy much more efficiently.

\section{Experiments}\label{sec:experiments}

In our experiments, we aim to answer the following research questions:

\noindent \textbf{RQ1 (Accuracy):} Does \textsc{ARES} perform better than existing anomaly detection methods?\\
\textbf{RQ2 (Ablation Study):} How do different components and design choices of \textsc{ARES} contribute to its performance?

\subsection{Datasets and Experimental Setup}

\begin{table}
\centering
\begin{tabular}{ccccc}
\toprule
\textbf{Dataset} & \#\textbf{Dim} & \#\textbf{Classes} & \#\textbf{Samples} & \textbf{Description} \\
\midrule 
SNSR \cite{SNSR}& $48$ & Multi-class & $58,509$ & Electric current signals \\
MNIST \cite{MNIST} & $784$ & Multi-class & $70,000$ & $0$-$9$ digit images \\
FMNIST \cite{xiao2017} & $784$ & Multi-class & $70,000$ & Fashion article images \\
OTTO  \cite{OTTO} & $93$ & Multi-class & $61,878$ & E-commerce types \\
MI-F \cite{MI} & $58$ & Single-class & $25,286$ & CNC milling defects \\
MI-V \cite{MI} & $58$ & Single-class & $23,125$ & CNC milling defects \\
EOPT \cite{EOPT} & $20$  & Single-class& $90,515$ & Storage system failures \\
\bottomrule
\end{tabular}
\caption{Name and descriptions of the datasets used in experiments, including the number of samples and dimensions.}
\label{datasets1}
\end{table}

Table \ref{datasets1} shows the datasets we use in experiments. The single-class datasets consist of ground truth normal-vs-anomaly labels, as opposed to the multiple class labels in multi-class datasets. In the latter, we distinguish between the 'one-class normality' setup, in which one class label is used as the normal class and all other classes are anomalous. Alternatively, in the 'multi-class normality' setup, one class is anomalous and all other classes are normal. For a dataset with $N$ classes, there are $N$ possible arrangements of normal and anomaly classes. We train separate models for each arrangement and find their average score for the final result. We use the Area-Under-Curve (AUC) metric to measure performance as it does not require an anomaly score threshold to be set.

As anomaly scores are calculated for each test sample independently of each other, the proportion of anomalies in the test set has no impact on the anomaly score assigned to any given sample. Therefore, we are able to use a normal:anomaly ratio of 50:50 in our experiments for the sake of simplicity and an unbiased AUC metric. Besides the normal sampels in the test set, the remaining normal samples are split $80$:$20$ into training and validation sets for all models. Full implementation details can be found in the supplementary material.

\subsection{Baselines}

We test the performance of \textsc{ARES} against a range of baselines. We use the sci-kit learn implementations of \textsc{LOF} (in the input space) \cite{Breunig2000}, \textsc{IForest} \cite{liu2012isolation}, \textsc{PCA} \cite{shyu2003novel} and \textsc{OC-SVM} \cite{Chen2001}. Publicly available codes are used for \textsc{DAGMM} \cite{Bo2018}, \textsc{RAPP-SAP} and \textsc{RAPP-NAP} \cite{Kim2019}, and we use Pytorch to build the autoencoder (\textsc{AE} and \textsc{ARES}) and variational autoencoder (\textsc{VAE}). All experiments were conducted in Windows OS using an Nvidia GeForce RTX 2080 Ti GPU.

We do not tune hyper-parameters relating to the model architectures or training procedures for any method. The effect of variation in hyper-parameters is studied in the ablation study in Section \ref{sec:ablation} and the supplementary material instead. We set the number of neighbours $k=10$ for both the local density and local reconstruction score in our main experiments.

\subsection{RQ1 (Accuracy):}

Table \ref{results} shows the average AUC scores as a percentage (i.e. multiplied by 100). In the one-class normality setting, \textsc{ARES} significantly improves performance over the baselines in all multi-class datasets besides \texttt{FMNIST}. In the single-class datasets, this improvement is even greater, e.g. $+8\%$ lift for \texttt{MI-F} and \texttt{EOPT}. Compared with \textsc{AE}, we see that local adaptivity helps to detect true anomalies by correcting for the natural variation in reconstruction error in the latent space. This effect is even more pronounced in the multi-class normality setting, where \textsc{ARES} gives the best performance on all datasets. Here, the normal set is much more diverse and therefore local adaptivity is even more important. We show the standard deviations and additional significance test scores in the supplementary material,  which also show statistically significant ($p<0.01$) improvement.

\begin{table}
\centering
\setlength{\tabcolsep}{1pt}
\begin{tabular}{ccccccccccc}
\toprule
Dataset & \textsc{LOF} & \textsc{IForest} & \textsc{PCA} & \textsc{OC-SVM} & \textsc{SAP} & \textsc{NAP} & \textsc{DAGMM} & \textsc{VAE} & \textsc{AE} & \textsc{ARES} \\
\midrule
\multicolumn{11}{c}{One-class Normality} \\
\midrule
\texttt{\texttt{SNSR}} & $97.98$ & $89.16$ & $92.01$ & $95.85$ & $98.79$ & $98.74$ & $88.08$ & $89.49$ & $98.30$ & \ \ \ \ $\mathbf{98.83}$**\\
\texttt{MNIST} & $96.85$ & $85.44$ & $95.68$ & $90.35$ & $95.35$ & $97.25$ & $89.60$ & $91.73$ & $96.96$ & \ \ \ \ $\mathbf{97.89}$**\\
\texttt{FMNIST} & $91.35$ & $91.39$ & $90.13$ & $90.74$ & $89.66$ & $\mathbf{93.08}$ & $87.97$ & $77.59$ & $92.33$ & $91.63$ \\
\texttt{OTTO} & $84.76$ & $70.34$ & $80.09$ & $81.43$ & $81.61$ & $82.77$ & $68.02$ & $82.39$ & $85.26$ & \ \ \ \ $\mathbf{87.86}$**\\
\texttt{MI-F} & $59.79$ & $81.53$ & $55.07$ & $76.69$ & $81.78$ & $80.61$ & $82.23$ & $76.93$ & $71.19$ & \ $\mathbf{89.52}$\\
\texttt{MI-V} & $83.97$ & $84.35$ & $87.32$ & $83.58$ & $88.24$ & $89.35$ & $75.45$ & $89.03$ & $90.75$ & \ $\mathbf{93.94}$\\
\texttt{EOPT} & $55.01$ & $61.61$ & $54.72$ & $59.66$ & $59.87$ & $61.69$ & $60.63$ & $68.08$ & $59.85$ & \ $\mathbf{68.43}$\\
\midrule
\multicolumn{11}{c}{Multi-class Normality} \\
\midrule
\texttt{SNSR} & $60.74$ & $52.70$ & $52.94$ & $52.79$ & $57.52$ & $58.32$ & $54.77$ & $61.36$ & $57.28$ &  \ \ \ \ $\mathbf{69.78}$**\\
\texttt{MNIST} & $77.40$ & $56.49$ & $70.41$ & $58.56$ & $84.76$ & $86.38$ & $54.24$ & $84.12$ & $80.04$ & \ \ \ \ $\mathbf{93.25}$**\\
\texttt{FMNIST} & $71.50$ & $64.75$ & $66.92$ & $60.27$ & $68.50$ & $72.09$ & $57.56$ & $71.18$ & $71.03$ & \ $\mathbf{72.49}$ \\
\texttt{OTTO}  & $63.01$ & $54.14$ & $58.27$ & $62.96$ & $57.47$ & $63.44$ & $58.96$ & $61.88$ & $59.59$ & \ $\mathbf{63.54}$ \\
\bottomrule
\end{tabular}
\caption{Mean AUC scores for each datasets and normality setting. The best scores are highlighted in bold and we mark the most significant improvements over \textsc{AE} ($p<0.01$) with **. Further tests are in the supplementary material.}
\label{results}
\end{table}

The neighbours of a normal sample with high reconstruction error tend to be have high reconstruction errors themselves. By basing the anomaly scoring on their relative difference, \textsc{ARES} uses this to better detect truly anomalous samples. Furthermore, \textsc{ARES} also uses the local density of the point, which depends purely on its distance to the training samples in the latent space. This is important as there may be some anomalies with such low reconstruction error that comparison with neighbours does not alone indicate anomalousness (for example the weekend samples mentioned earlier). These samples could be expected to be occupy very sparse regions of the latent space due to their significant deviation from the normal set, which means they can be better detected through the local density score. By combining these two scores, we are able to detect a wider range of anomalous data than either could individually.

\subsection{RQ2 (Ablation Study):}\label{sec:ablation}

\paragraph{Density Estimation Method}
Table \ref{lds_results} shows the performance of \textsc{ARES} with other density estimation methods. \textsc{KNN} is the distance to the $k^{th}$ nearest neighbour in the latent space with $k=20$. \textsc{GD} is the distance of a point to the closest of $N$ Gaussian distributions fit to the latent encodings of samples for each of the classes in the training set ($N$ = 1 in the one-class normality case). \textsc{NF} is the likelihood under a RealNVP normalizing flow \cite{Papamakarios2017}.

\begin{table}
\centering
\setlength{\tabcolsep}{20pt}
\begin{tabular}{ccccc}
\toprule
Dataset &\textsc{LOF} & \textsc{KNN} & \textsc{GD} & \textsc{NF} \\
\midrule
\multicolumn{5}{c}{One-class Normality} \\
\midrule
\texttt{SNSR} & $\mathbf{98.83}$ & $98.66$ & $95.68$ & $98.29$\\
\texttt{MNIST} & $\mathbf{97.89}$ & $95.24$ & $87.29$ & $97.30$\\
\texttt{FMNIST} & $91.63$ & $\mathbf{92.94}$ & $90.26$ & $92.14$\\
\texttt{OTTO} & $\mathbf{87.76}$ & $83.14$ & $81.27$ & $86.74$\\
\texttt{MI-F} & $\mathbf{89.52}$ & $76.12$ & $80.41$ & $84.47$\\
\texttt{MI-V} & $\mathbf{93.94}$ & $91.60$ & $79.61$ & $92.89$\\
\texttt{EOPT} & $\mathbf{68.43}$ & $67.63$ & $63.35$ & $61.07$\\
\midrule
\multicolumn{5}{c}{Multi-class Normality} \\
\midrule
\texttt{SNSR} & $\mathbf{69.78}$ & $67.42$ & $61.86$ & $60.63$ \\
\texttt{MNIST} & $\mathbf{93.25}$ & $92.92$ & $91.14$ & $86.25$\\
\texttt{FMNIST} & $72.49$ & $72.77$ & $\mathbf{73.02}$ & $70.48$\\
\texttt{OTTO} & $63.54$ & $61.71$ & $\mathbf{66.30}$ & $63.67$\\
\bottomrule
\end{tabular}
\caption{Mean AUC scores for each choice of local density score. The best scores are highlighted in bold.}
\label{lds_results}
\end{table}

\textsc{LOF} performs best overall, closely followed by \textsc{KNN}. \textsc{GD} performs poorly in one-class experiments, however it is better in multi-class normality experiments and even the best for \texttt{FMNIST} and \texttt{OTTO}. This could be as the use of multiple distributions provides more flexbility. \textsc{NF} is noticeably worse; previous studies have found that normalizing flows are not well-suited to detect out-of-distribution data \cite{kirichenko2020normalizing}.

In the supplementary material, we further test these density estimation methods by varying their hyper-parameters and find \textsc{LOF} to still come out best. Further supplementary experiments show that the local density score generally performs better than the local reconstruction score in the multi-class normality case and vice versa in the one-class case. Combining them, as in \textsc{ARES}, generally gives the best performance overall across different latent embedding sizes.

\paragraph{Robustness to Training Contamination}

In practice, it is likely that a small proportion of anomalies `contaminate' the training set. In this section, we study the effect of different levels of contamination on ARES, defined as $n\%$ of the total number of samples in the training set. Table \ref{results2} shows that increasing $n$ worsens performance overall. With more anomalies in the training set, it is more likely that anomalies are found in the nearest neighbour set of more test points, which skews both the average neighbourhood reconstruction error as well as their density in the latent space and degrades performance.

\begin{table}
\centering
\setlength{\tabcolsep}{15pt}
\begin{tabular}{cccccc}
\toprule
\multicolumn{6}{c}{Neighbourhood Size} \\
\midrule
$n$ & $10$ & $50$ & $100$ & $200$ & $500$\\
\midrule
\multicolumn{6}{c}{One-class Normality} \\
\midrule
$0$ & $\mathbf{97.89}$ & $97.85$ & $97.80$ & $97.73$ & $97.56$ \\
$0.5$ & $95.77$ & $\mathbf{96.08}$ & $96.06$ & $95.95$ & $95.72$ \\
$1$ & $82.50$ & $94.45$ & $\mathbf{95.97}$ & $95.81$ & $95.42$ \\
$2$ & $90.65$ & $92.88$ & $93.29$ & $\mathbf{93.33}$ & $92.99$ \\
$3$ & $88.41$ & $91.35$ & $91.92$ & $\mathbf{92.07}$ & $91.86$ \\
$5$ & $84.91$ & $87.85$ & $89.20$ & $90.26$ & $\mathbf{90.65}$ \\
$10$ & $79.69$ & $81.51$ & $83.16$ & $84.88$ & $\mathbf{86.82}$ \\
\midrule
\multicolumn{6}{c}{Multi-class Normality} \\
\midrule
$0$ & $\mathbf{93.25}$ & $92.71$ & $92.08$ & $91.31$ & $89.68$\\
$0.5$ & $68.38$ & $76.07$ & $78.36$ & $80.75$ & $\mathbf{81.42}$\\
$1$ & $52.38$ & $58.42$ & $67.62$ & $65.75$ & $\mathbf{68.91}$\\
$2$ & $59.50$ & $63.13$ & $65.71$ & $67.98$ & $\mathbf{69.68}$\\
$3$ & $57.09$ & $59.00$ & $60.78$ & $62.89$ & $\mathbf{65.35}$\\
$5$ & $50.86$ & $51.48$ & $52.40$ & $53.47$ & $\mathbf{55.78}$\\
$10$ & $\mathbf{52.50}$ & $50.80$ & $50.87$ & $51.14$ & $52.11$\\
\bottomrule
\end{tabular}
\caption{Mean AUC scores in \texttt{MNIST} with training set anomaly contamination ($n$\%) and different neighbourhood sizes. The best scores are highlighted in bold.}
\label{results2}
\end{table}

We find that ARES is more robust to training set contamination with a higher setting of the number of neighbours ($k$). In the one-class setup, we see that ARES performs better with higher values of $k$ as the proportion of anomalies increases. By using more neighbours, the effect of any individual anomalies on the overall neighbourhood error is reduced, which helps to maintain better performance. This effect is even more stronger in the multi-class normality setup. The highest neighbour count of $k=500$ gives the best performance in all cases except for $n=0\%$ and $10\%$. As the multi-class training sets are much larger than their one-class normality counterparts, $n\%$ corresponds to a much larger number of anomalous contaminants, which explains their greater effect for a given $k$.

\section{Conclusion}

Autoencoders are extremely popular deep learning models used for anomaly detection through their reconstruction error. We have shown that the assumption made by the standard reconstruction error score, that reconstruction errors are identically distributed for all normal samples, is unsuitable for real datasets. We empirically show that there is a heteroscedastic relationship between latent space characteristics and reconstruction error, which demonstrates why adaptivity to local latent information is important for anomaly scoring. As such, we have developed a novel approach to anomaly scoring which adaptively evaluates the anomalousness of a samples reconstruction error, as well as its density in the latent space, relative to those of its nearest neighbours. We show that our approach results in significant performance improvements over the standard approach, as well as other prominent baselines, across a range of real datasets.

\section*{Acknowledgements}

This work was supported in part by NUS ODPRT Grant R252-000-A81-133.

\bibliographystyle{splncs04}
\bibliography{ref}

\end{document}


\title{Supplementary Material: Appendix} 
%
%
%
%
%
%
\author{}
\institute{}
\maketitle              
%






\title{Supplementary Material: Appendix}

\section*{Experiments}

\subsection*{Model Architecture and Training}

The autoencoder models for the two image datasets had seven hidden layers in both the encoder and decoder, for $14$ layers total, while the tabular datasets had six in each ($12$ total). All linear layers were followed by a Leaky ReLU layer and Batch Normalization except for the final output layer in the decoder. For the image datasets, the hidden layer sizes for the encoder were as follows: [$784,600,500,400,300,200,100,20$] and the same in reverse order for the decoder. For tabular datasets, the bottleneck size was set equal to the number of \textsc{PCA} principal components needed to explain $90\%$ of the data, and the hidden layer sizes  decrease (increase) linearly in the encoder (decoder) accordingly. All autoencoder models (including \textsc{VAE}) models followed this structure for the sake of consistency. All models were trained for $350$ epochs with early stopping activated if the loss function, mean squared error, did not achieve a new minimum for $20$ consecutive epochs. The batch size was set at $250$ samples. 

\subsection*{Standard Deviation in AUC Score}

Table \ref{sd1} shows the standard deviations of the AUC scores over the $N$ trials for each dataset and settings, where $N$ is the number of classes in the dataset. Naturally, as each setup has a different set of normal and anomalous classes, the scores vary greatly. Therefore we show the ificance values for performance improvement in the next section.

\begin{table*}
\centering
\setlength{\tabcolsep}{5pt}
\begin{tabular}{c|cccccccc}
\toprule
Dataset & \textsc{PCA} & \textsc{LOF} & \textsc{OC-SVM} & \textsc{AE} & \textsc{SAP} & \textsc{NAP} & \textsc{DAGMM} & \textsc{ARES} \\
\midrule
\multicolumn{9}{c}{Multi-Class Normality} \\
\midrule
\texttt{SNSR}    & 7.91 & 18.47 & 19.16 & 17.25 & 18.52 & 20.19 & 13.15 & 16.73    \\
\texttt{MNIST}   & 16.48 & 14.91 & 16.65 & 22.64 & 13.57 & 12.54 & 14.30 & 7.77     \\
\texttt{FMNIST}  & 16.12 & 17.01 & 17.10 & 17.43 & 17.48 & 15.20 & 23.79 & 16.23 \\
\texttt{OTTO}    & 15.24 & 22.76 & 19.58 & 17.62 & 17.88 & 20.32 & 15.64 & 18.09    \\
\midrule
\multicolumn{9}{c}{One-Class Normality} \\
\midrule
\texttt{SNSR}    & 5.41 & 1.12 & 2.54 & 1.32 & 0.66 & 0.67 & 6.38 & 0.94     \\
\texttt{MNIST}   & 4.11 & 2.72 & 6.59 & 2.44 & 2.82 & 1.90 & 5.76 & 1.75   \\
\texttt{FMNIST}  &  6.25 & 4.77 & 5.74 & 5.07 & 6.12 & 4.84 & 8.62 & 4.04     \\
\texttt{OTTO}    & 8.64 &  6.46 &  7.17 & 6.15 & 7.63 & 7.27 & 8.44 & 4.98    \\

\bottomrule
\end{tabular}
\caption{Standard deviation of AUC scores over the different runs.}
\label{sd1}
\end{table*}

\subsection*{One-sided Wilcoxon Significance Test}

\begin{table}
\caption{p values from the one-sided Wilcoxon signed-rank test. Values lower than 0.05 and 0.01 are marked with * and ** respectively.}
\centering
\begin{tabular}{p{3cm}p{2cm}p{2cm}}
\toprule 
$p$ value & AE & RAPP-NAP  \\
\midrule
\multicolumn{3}{c}{One-Class Normality}   \\
\midrule
\texttt{SNSR} & 0.0063** & 0.2969 \\
\texttt{MNIST} & 0.0083** & 0.0234* \\
\texttt{FMNIST} & 0.9303 & 0.9937 \\
\texttt{OTTO} & 0.0038** & 0.0038** \\
\midrule
\multicolumn{3}{c}{Multi-Class Normality} \\
\midrule
\texttt{SNSR} & 0.0017** & 0.0029** \\
\texttt{MNIST} & 0.0035** & 0.0035** \\
\texttt{FMNIST} & 0.2538 & 0.6006 \\
\texttt{OTTO} & 0.0693 & 0.6165 \\
\midrule
\bottomrule
\end{tabular}
\label{wilcoxon_test}
\end{table}

In Table \ref{wilcoxon_test}, we show the $p$-values of the one-sided Wilcoxon signed-rank test [1]. We test our method against both the \textsc{AE}, to see the effect of local adaptivity, as well as \textsc{RAPP-NAP} performance, as it was generally the next best-performing method. A lower $p$-value indicates more confidence that our method performs better over the corresponding baseline method. The binary-class datasets are not shown as they only had one experimental setup, i.e. one AUC score to measure. The $p$-values for the \texttt{SNSR}, \texttt{MNIST} and \texttt{OTTO} datasets in particular are extremely low. The performance is more comparable between \textsc{ARES} and the baselines for \texttt{FMNIST}, hence the larger $p$-values. The median $p$-value of \textsc{ARES} against AE is $0.0073$ and for \textsc{RAPP-NAP} it is $0.016$.

\subsection*{Runtime}

Table \ref{runtimes} shows the runtimes of the autoencoder training, the regular \textsc{AE} anomaly scoring and our \textsc{ARES} scoring. We see that, despite noticeably longer computational time for our method at test time, it is very insignificant in relation to the overall pipeline of the model including training time.

\begin{table}
    \centering
     \setlength{\tabcolsep}{15pt}
    \begin{tabular}{cccc}
    \toprule
    Time & Training & \textsc{AE} score & \textsc{ARES} score \\
    \midrule
    One-Class & 1.47444 & 0.00014 & 0.01721 \\
    Multi-Class & 24.29161 & 0.00037 & 0.87681 \\
    \bottomrule
    \end{tabular}
    \caption{Runtimes (minutes) of training and testing for \textsc{AE} and \textsc{ARES} for comparison for each normality setting. Runtimes are calculated for the \texttt{MNIST} dataset and averaged over each of the trials for each modality setup.}
    \label{runtimes}
\end{table}

\subsection*{Analysis} 
\begin{figure}
     \centering
     \includegraphics[width=0.6\textwidth]{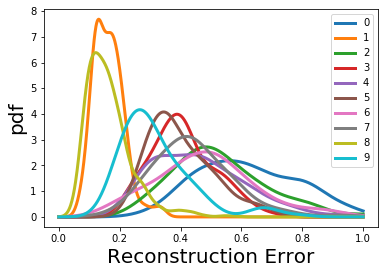}
     \caption{Probability density functions of reconstruction errors of each class (class 8 is normal and others are anomalous.}
     \label{individual1}
\end{figure}

To understand why local adaptivity improves anomaly detection performance, we study one particular problem setup: \texttt{MNIST} samples of class 8 are normal and all other classes are all anomalies. Figure \ref{individual1} shows the reconstruction error of the test samples of all classes in this setup. We see that reconstruction errors of the class 8 samples are generally lower than those from other classes except for class 1. This means that, based on reconstruction error alone, many anomalies from class 1 will be falsely classified as normal. 

\begin{figure}
     \centering
     \includegraphics[width=0.6\textwidth]{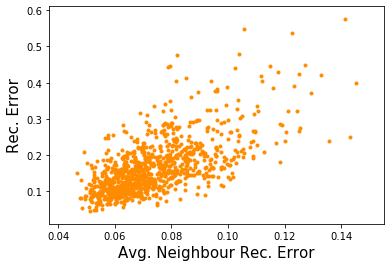}
     \caption{Reconstruction error of normal (8) test samples versus the average reconstruction error of their training set nearest neighbours.}
     \label{individual3}
\end{figure}

\textsc{ARES} compares a points reconstruction error against its nearest neighbours which provides local context for more accurate detection. As shown in Figure \ref{individual3}, the neighbours of a poorly reconstructed normal test point tend to be the more poorly reconstructed training points, perhaps because these training points also contained some kind of abnormality. With this, the abnormality of these normal test points is less outlying within the context of the training points around them. This context is neglected in the standard scoring approach, and so is likely to detect false positives for these kinds of samples, unlike in our method which accounts for this context. 

 \begin{figure}
     \centering
     \includegraphics[width=0.6\textwidth]{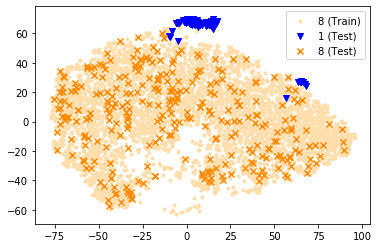}
     \caption{t-SNE embeddings of class 1 test samples (blue), class 8 test samples (dark orange) and class 8 training samples (light orange).}
     \label{individual2}
\end{figure}

Furthermore, \textsc{ARES} also uses the local density of the point, which depends not on its reconstruction error but purely on its distance to the training samples in the latent space. In Figure \ref{individual2}, we can see that the vast majority of anomalous class 1 samples (in blue) are far away from their nearest neighbours in the training set than the normal class 8 test samples (in dark orange). By combining these two scores, we are able to detect such anomalies even if the reconstruction error alone cannot.

\subsubsection*{Scaling factor}

In Table \ref{alpha_results}, we see the average AUC score found when using different values for the scaling factor $\alpha$. A lower or higher value of the scaling factor gives more weight to the local reconstruction or density score respectively.

Generally, we see that a lower value of $\alpha$ is better in the one-class normality case, while higher is better for the multi-class normality case, meaning the local density score is more beneficial when the diversity in the normal set is larger. This corroborates the results seen in Table \ref{components}, as the local reconstruction score is relatively better than the local density score in the one-class normality case and vice versa in the multi-class normality case.

\begin{table}
    \centering
    \setlength{\tabcolsep}{8pt}
    \begin{tabular}{c|cc|cc}
    \toprule
    \multicolumn{1}{c}{} & \multicolumn{2}{c}{One-Class Normality} & \multicolumn{2}{c}{Multi-Class Normality} \\
    \multicolumn{1}{c}{$\alpha$} & \texttt{MNIST} & \multicolumn{1}{c}{\texttt{FMNIST}} & \multicolumn{1}{c}{\texttt{MNIST}} & \texttt{FMNIST} \\ \midrule
    $0.1$ & $97.51$ & $91.52$ & $88.08$ & $70.84$ \\ 
    $0.25$ & $97.76$ & $\mathbf{91.70}$ & $91.53$ & $71.97$ \\ 
    $0.5$ & $\mathbf{97.89}$ & $91.63$ & $93.25$ & $\mathbf{72.49}$ \\ 
    $1$ & $97.72$ & $91.16$ & $\mathbf{93.91}$ & $\mathbf{72.49}$ \\ 
    $2$ & $97.00$ & $90.25$ & $93.90$ & $72.17$ \\ 
    \bottomrule
    \end{tabular}
    \caption{Average AUC score (\%) of ARES with different settings of scaling factor $\alpha$}
    \label{alpha_results}
\end{table}

\subsubsection*{Score Comparison}

Table \ref{components} shows the performance of the local density and local reconstruction scores separately. We see that the local reconstruction score tends to do better in the one-class normality experiments while the local density score is better for multi-class normality experiments. However, combining both of them as in ARES gives the best performance in most cases overall.

\begin{table}[h!]

\centering
\caption{AUC score of the two components of ARES in isolation.}
\begin{tabular}{{p{3cm}p{2cm}p{2cm}p{2cm}}}
\toprule
Dataset & $d(\mathbf{x})$ & $r(\mathbf{x})$ & \textsc{ARES}\\
\midrule
\multicolumn{4}{c}{One-Class Normality} \\
\midrule
\texttt{SNSR} & $96.24$ & $98.49$ & $\mathbf{98.83}$\\
\texttt{MNIST} & $91.59$ & $97.21$ & $\mathbf{97.89}$ \\
\texttt{FMNIST} &$86.97$ & $91.17$ & $\mathbf{91.63}$ \\
\texttt{OTTO} & $77.17$ & $\mathbf{88.57}$ & $87.86$ \\
\texttt{MI-F} & $87.08$ & $68.69$ & $\mathbf{89.52}$ \\
\texttt{MI-V} & $87.93$ & $\mathbf{94.73}$ & $93.94$ \\
\texttt{EOPT} & $67.88$ & $60.13$ & $\mathbf{68.43}$\\
\midrule
\multicolumn{4}{c}{Multi-Class Normality} \\
\midrule
\texttt{SNSR} & $\textbf{70.46}$ & $62.66$ & $69.78$\\
\texttt{MNIST} & $93.23$ & $81.96$ & $\mathbf{93.25}$ \\
\texttt{FMNIST} & $71.05$ & $68.97$ & $\mathbf{72.49}$ \\ 
\texttt{OTTO} & $62.54$ & $60.84$ & $\mathbf{63.54}$ \\
\bottomrule
\end{tabular}
\label{components}
\end{table}

\newpage
\subsubsection*{Density Estimation Method}

\begin{table}
    \setlength{\tabcolsep}{5pt}
    \centering
    \begin{tabular}{cl|cccccccc}
    \toprule
    Dataset & & \texttt{SNSR} & \texttt{MNIST} & \texttt{FMNIST} & \texttt{OTTO} & \texttt{MI-F} & \texttt{MI-V} & \texttt{EOPT}  \\ 
    \midrule
    \multicolumn{10}{c}{One-Class Normality} \\ 
    \midrule
    LOF & $k=10$ & $\mathbf{98.83}$ & $\mathbf{97.89}$ & $91.63$ & $87.76$ & $\mathbf{89.52}$ & $\mathbf{93.94}$ & $68.43$ \\ 
    & $k = 40$ & $98.75$ & $97.85$ & $92.53$ & $88.04$ & $79.31$ & $91.99$ & $62.28$  \\ 
    & $k = 100$ & $98.55$ & $97.80$ & $\mathbf{93.02}$ & $\mathbf{88.08}$ & $69.69$ & $89.98$ & $60.80$ \\ 
    KNN & $k=5$ & $98.65$ & $95.53$ & $92.47$ & $83.99$ & $77.26$ & $93.55$ & $\mathbf{79.45}$ \\ 
    & $k=20$ & 98.66 & 95.24 & 92.94 & 83.14 & 76.12 & 91.60 & 67.63 \\ 
    & $k=40$ & 98.56 & 94.93 & 92.99 & 82.76 & 74.24 & 90.508 & 64.91  \\ 
    GD & Euclidean & 97.21 & 72.74 & 79.47 & 82.08 & 80.80 & 75.77 & 61.11 \\ 
    & Mahalanobis & 95.68 & 87.29 & 90.26 & 81.27 & 80.41 & 79.61 & 63.35 \\ 
    \textsc{NF} & & 98.29 & 97.30 & 92.14 & 86.74 & 84.47 & 92.89 & 61.07 \\ 
    \midrule
    \multicolumn{10}{c}{Multi-Class Normality} \\
    \midrule
    LOF & $k=10$ & $\mathbf{69.78}$ & $93.25$ & $72.49$ & $63.54$ & ~ & ~ & ~ \\ 
    & $k = 40$ & $67.65$ & $92.90$ & $72.74$ & $63.80$ & ~ & ~ & ~ & ~ \\ 
    & $k = 100$ & $66.36$ & $92.08$ & $72.48$ & $\mathbf{64.73}$ & ~ & ~ & ~ & ~ \\ 
    KNN & $k=5$ & $68.63$ & $\mathbf{93.68}$ & $\mathbf{73.52}$ & $61.76$ & ~ & ~ & ~ \\ 
    & $k=20$ & 67.42 & 92.92 & 72.77 & 61.74 & ~ & ~ & ~ & ~ \\ 
    & $k=40$ & 66.50 & 92.06 & 72.28 & 61.74 & ~ & ~ & ~ & ~ \\ 
    GD & Euclidean & 56.55 & 74.40 & 66.00 & 59.38 & ~ & ~ & ~ \\ 
    & Mahalanobis & 61.86 & 91.14 & 73.02 & 66.30 & ~ & ~ & ~ & ~ \\ 
    \textsc{NF} & & 60.63 & 86.25 & 70.48 & 63.67 & ~ & ~ & ~ & ~ \\ 
    \bottomrule
    \end{tabular}
    \caption{Mean AUC scores for each choice of local density score. The best scores are highlighted in bold.}
    \label{lds_results}
\end{table}

Table \ref{lds_results} shows the performance of \textsc{ARES} with other density estimation methods. \textsc{KNN} is the distance to the $k^{th}$ nearest neighbour in the latent space. \textsc{GD} is the distance of a point to a Gaussian distribution fit to the training set latent encodings or the closest of $N$ in the multimodal case. \textsc{GD} performs poorly in unimodal experiments, however it is better in multimodal experiments and even the best for \texttt{FMNIST} and \texttt{OTTO}. This could be as there are $N$ separate distributions used for each of the $N$ normal classes in this case. Within \textsc{GD}, we test the use of the Mahalanobis distance versus the Euclidean distance, and we see the former outperform the latter in most cases. \textsc{NF} is the likelihood under a RealNVP normalizing flow [2]. Other flows were tested but did not give stable results. It is noticeably worse than the other methods; previous studies have found that normalizing flows are not well-suited to detect out-of-distribution data.

\newpage

\subsubsection*{Embedding Size}

We also study the effect of latent size, or the dimensionality of latent encodings, on the performance of our approach as well as the following variants:

\begin{itemize}
    \item \textbf{AE:} The standard AE method without local adaptivity.
    \item \textbf{ARES-G:} AE reconstruction score combined with local density score.
    \item \textbf{L-R:} Local reconstruction score only.
    \item \textbf{L-D:} Local density score only.
    \item \textbf{ARES}: Combined local reconstruction and density score.
\end{itemize}

\begin{figure}
    \centering
    \includegraphics[width=0.8\linewidth]{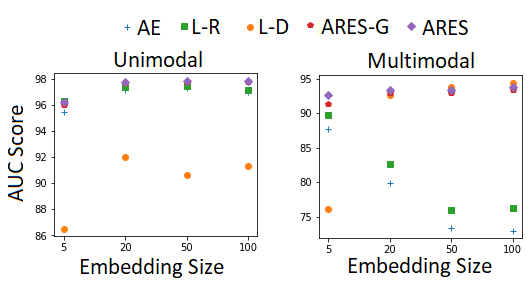}
    \caption{Performance of various component methods across a range of embedding sizes.}
    \label{ablation_embedding}
\end{figure}

In Figure \ref{ablation_embedding}, we see this effect for the MNIST dataset in particular. We see that embedding size has little effect on reconstruction scores in the one-class normality setting but reduces performances significantly in the multi-class normality setting. In both settings, however, the density scores mostly improve with increasing embedding size. Overall, we see that combining the local reconstruction score with the density score gives the best performance across the board; it is more robust to variation in hyper-parameter choices.

\section*{References}
1. Demsar, J.: Statistical comparisons of classifiers over multiple data sets. The Journal
of Machine Learning Research 7, 1–30 (2006) \\
2. Papamakarios, G., Pavlakou, T., Murray, I.: Masked autoregressive flow for density
estimation. In: NeurIPS. pp. 2338–2347 (2017)

